\documentclass[letterpaper, 10 pt, conference]{ieeeconf}  

\IEEEoverridecommandlockouts                              

\usepackage{siunitx}
\usepackage{amsmath}
\usepackage{amssymb}
\usepackage{amsfonts}
\usepackage{nicefrac}
\usepackage{soul}
\usepackage{hyperref}
\usepackage{float}

\usepackage{todonotes}

\title{\LARGE \bf
Physics-based Simulation of Continuous-Wave LIDAR\\
for Localization, Calibration and Tracking
}

\author{Eric Heiden, Ziang Liu, Ragesh K. Ramachandran, Gaurav S. Sukhatme
\thanks{The authors are with the Department of Computer Science,
        University of Southern California, Los Angeles, USA.
        {\tt\small heiden@usc.edu}}%
}

\usepackage{xspace}
\newcommand*{\eg}{e.g.\@\xspace}

\newcommand*{\etc}{etc.\@\xspace}

\begin{document}

\maketitle
\thispagestyle{empty}
\pagestyle{empty}

\begin{abstract}

Light Detection and Ranging (LIDAR) sensors play an important role in the perception stack of autonomous robots, supplying mapping and localization pipelines with depth measurements of the environment.
While their accuracy outperforms other types of depth sensors, such as stereo or time-of-flight cameras, the accurate modeling of LIDAR sensors requires laborious manual calibration that typically does not take into account the interaction of laser light with different surface types, incidence angles and other phenomena that significantly influence measurements. In this work, we introduce a physically plausible model of a 2D continuous-wave LIDAR that accounts for the surface-light interactions and simulates the measurement process in the Hokuyo URG-04LX LIDAR. Through automatic differentiation, we employ gradient-based optimization to estimate model parameters from real sensor measurements.

\end{abstract}

\section{Introduction}

Light detection and ranging (LIDAR) sensors, also known as laser range finders (LRF), are active depth sensors that transmit and receive laser light to measure the distance to objects that reflect light back to the sensor. Since the advent of compact planar LIDARs, such as Sick LMS 200 or Hokuyo URG-04LX, they have become an important depth sensor in the perception stack of autonomous mobile robots. 

LRF sensors exhibit unique noise characteristics. Their measurements are influenced by the distance to the object, the angle of incidence, the surface properties of objects (\eg reflectivity and color), the environment temperature, and many other factors, which have been studied extensively~\cite{ye2002char, kneip2009urg}. The goal of such characterizations is to derive an empirical sensor model ``top-down'': given LIDAR measurements under varying conditions, such as surface types, inclination angles, \etc, they produce a probabilistic model of noise characteristics for a particular sensor. This statistical model then further informs uncertainty-aware localization and mapping algorithms.

In this work, we take a different approach: starting from first principles on how laser light interacts with surfaces under various conditions, we implement a physically plausible simulation that recovers various effects encountered with physical LIDARs, such as spurious measurements, reflected and refracted rays. We simulate the measurement process of continuous-wave LIDARs where the phase shift between two amplitude-modulated laser light waves is calculated to measure the distance. Using automatic differentiation, we compute the gradients of all parameters in our simulation with respect to the simulated measurements and apply gradient-based optimizers to find the simulation parameters that most closely match the true observations.

To the best of our knowledge, our approach is the first physically plausible simulation that captures the measurement process of a continuous-wave LIDAR. Our real-world experiments show that our proposed model can accurately reproduce light-surface phenomena, such as refraction and specular reflection, that have not been considered in most previous LIDAR models.


\begin{figure}[t]
    \centering
    \includegraphics[width=\columnwidth]{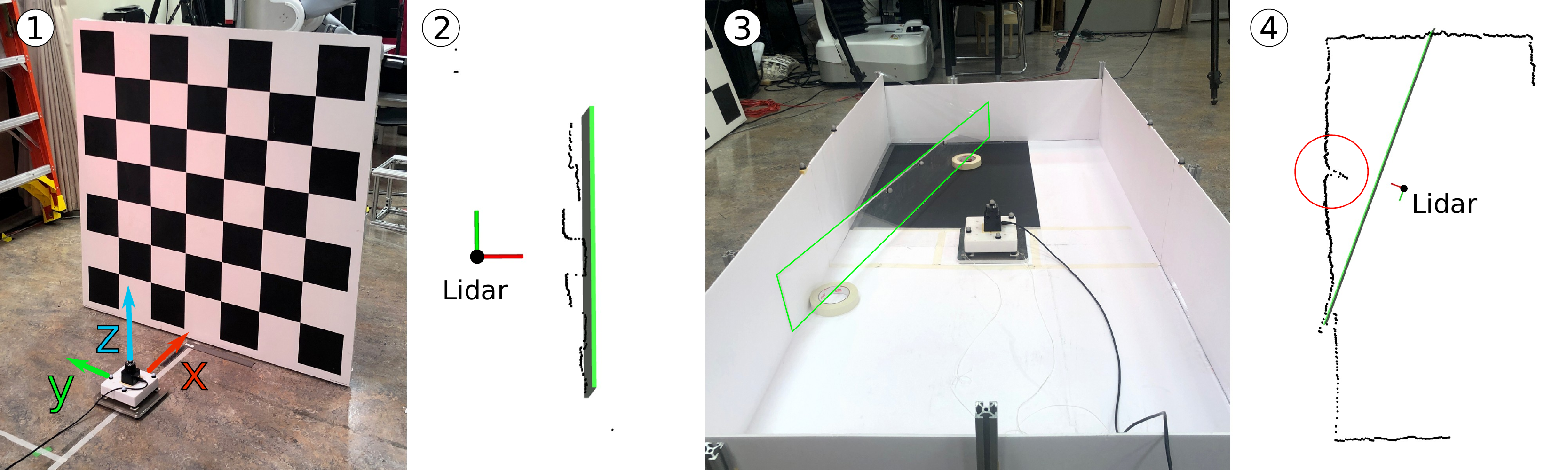}
    \caption{Various effects encountered on the Hokuyo URG-04LX laser scanner.
    (1) LIDAR at a distance of \SI{0.5}{\meter} in front of a checkerboard. (2) Measured point cloud obtained from the checkerboard with notable staircase pattern. (3) LIDAR in the cuboid environment, used for the localization and tracking experiments, with a sheet made of transparent plastic (outlined in green). (4) Sensed point cloud where the plastic sheet is invisible, except for a distortion that appears perpendicular to the LIDAR.}
    \label{fig:effects}
\end{figure}

\begin{figure*}[t!]
    \centering
    \includegraphics[width=.9\textwidth]{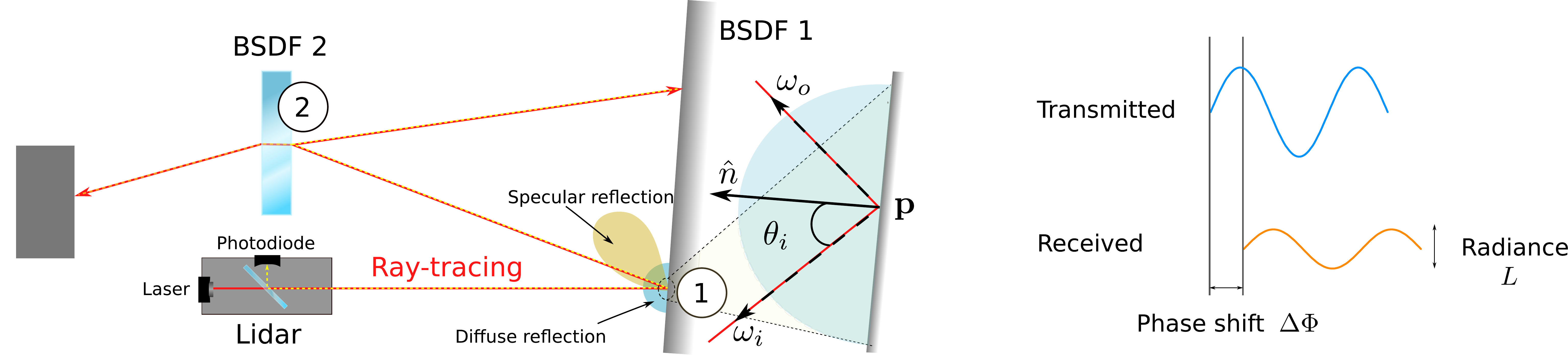}
    \caption{\textit{Left:} LIDAR principle where light emanating from a laser is captured by the photodiode on the other side of the half-transparent mirror. The laser ray is traced through the scene to compute the received radiance $L$ (\autoref{eqn: output radiance}) by considering various surface interactions through bidirectional scattering distribution functions (BSDF). BSDF (1) shows the uniform scattering of light due to blackbody radiation, and diffusive and specular reflection spread due to the microfacets in the region around the intersection point $\mathbf{p}$ with surface normal $\hat{n}$. The magnified view of (1) depicts the vectors used for the formulation of BRDFs. BSDF (2) transmits and refracts a part of the light, while reflecting the other part (similar to glass). 
    \textit{Right:} attenuation in the LIDAR received wave compared to the transmitted wave. The phase shift of the received wave resulting from the light's travel time from the laser to an object and back to the photodiode is used to compute the distance (cf.~\autoref{eq:distance_phase}).}
    \label{fig:LIDAR_lte}
\end{figure*}

\section{Related Work}
\label{sec:related}

Detailed characterizations of 2D LIDAR sensors have been presented for the Hokuyo laser scanner URG-04LX in~\cite{kawata2008urg, okubo2009urg, kneip2009urg, breda2012experimental}. The focus of these works is on experimentally analyzing how the LIDAR measurements are influenced by object color, material, incidence angle, etc. Our goal, in contrast, is to model the physics behind light-surface interaction and the LIDAR's measurement process to allow for the automatic inference of the conditions that caused the observed measurements.
A detailed analysis of the physical properties of LIDARs is given in~\cite{rosenberger2018analysis}, where the authors provide suggestions on how certain real-world effects can be modeled in simulation.

LIDARs and other active range finders are typically modelled in robot simulators to allow for the development of localization and mapping algorithms. Instead of accounting for the intricacies unique to these sensors -- as done in the works on characterization -- most general-purpose simulators for robotics, such as Gazebo~\cite{koenig2004gazebo} and MuJoCo~\cite{todorov2012mujoco}, typically implement z-buffering or ray-casting to compute the depth and perturb it by zero-mean Gaussian noise, leading to a significant gap between simulation and reality~\cite{durst2011need}. In this work, we aim for a more detailed simulation that takes into account the interaction of laser light with different surfaces, and many other physical phenomena, as described in Sec.~\ref{sec:model}. Similar to~\cite{gschwandtner2011blensor}, we implement ray tracing to account for refractive and reflective surfaces. Additionally, we model the sampling process taking place in continuous-wave LRFs, as well as the detailed light-surface interaction that considers transmitting materials, such as glass and plastic.

Using gradients of models and algorithms has a long history in engineering, \eg where gradients of ray-tracing equations have been derived to optimize the design of optical systems~\cite{feder1968diffrt}. With the advent of automatic differentiation frameworks, accurate gradients can be computed algorithmically, powering model-predictive optimal control algorithms based on robotic simulators~\cite{giftthaler2017automatic, peres2018lcp, heiden2019ids}, design optimization~\cite{feder1968diffrt}, and various other use cases~\cite{barthelemy1995autodiff, corliss2002autodiff}. In the context of simulating vision, prior work in computer graphics has focused on differentiable rendering systems~\cite{loper2014opendr, henderson18bmvc, li2018differentiable, liu2019soft, NimierDavidVicini2019Mitsuba2} that tackle the problem of inverse graphics. Overall, our framework follows a similar idea of constructing a physics-based model of the real system for which the parameters can be efficiently estimated using gradient-based optimization.





	

\section{LIDAR Model}
\label{sec:model}
In the following section, we describe our simulation of a continuous-wave LIDAR, starting from the physics of light-surface interaction to the particular measurement process in the considered sensor.

\subsection{Surface Interaction}
\label{sec:surface}
When the laser light hits an object at a point $\mathbf{p}$, depending on the surface and angle of inclination, the beam is reflected, scattered and transmitted.

In computer graphics, researchers and practitioners use the rendering equation~\cite{pharr2016pbrt} 
\begin{align}
\label{eqn: output radiance}
    L_0(\mathbf{p}, \omega_o) = \ &L_e(\mathbf{p}, \omega_o) \\
    \nonumber
    &+ \int_{\Omega} f(\mathbf{p}, \omega_o, \omega_i) L_i(\mathbf{p}, \omega_i) |\cos{\theta_i}|\,\text{d}\omega_i
\end{align}
which expresses the radiance of the reflected light ray from point $\mathbf{p}$ along the direction $\omega_o$, when a light ray of radiance $L_i(\mathbf{p}, \omega_i)$ tracking the direction $\omega_i$ collides with a medium at $\mathbf{p}$, integrated over all incoming light directions $\omega_i$ from the unit hemisphere $\Omega$. $\theta_i$ is the angle made by the incident light ray with the surface normal at $\mathbf{p}$. These quantities are depicted in \autoref{fig:LIDAR_lte} (left).

The term $L_e(\mathbf{p}, \omega_o)$ captures the effects of blackbody emissions from the material on which the light was incident. All substances having a temperature above absolute zero emit electromagnetic radiation. The radiance of such electromagnetic radiation with a wavelength $\lambda$ emitted by a blackbody at a particular temperature can be derived using Planck`s law \cite[Chapter 12.1.1]{pharr2016pbrt}.

The integral term in \autoref{eqn: output radiance} quantifies the amount of radiance imbibed by the reflected light ray along $\omega_o$ from the light rays incident at $\mathbf{p}$. This term is referred to as a bidirectional reflectance distribution function (BRDF).
Fresnel equations give the fraction of light reflected along $\omega_o$, denoted as $F_r(\omega_o)$. Thus, if we set the reflection distribution function $f(\mathbf{p}, \omega_o, \omega_i) = F_r(\omega_o) \frac{\delta(\omega_i - \omega_o)}{|\cos{\theta_i}|}$, then the integral reduces to $F_r(\omega_o)L_i(\mathbf{p}, \omega_i)$. Incidentally, $F_r(\omega_o)$ is a function of the refractive indices of the participating media and can be computed in closed form~\cite[Chapter 8.2.1]{pharr2016pbrt}.

Apart from being specularly reflected, light can be scattered and diffused by the interacting surface. Diffuse reflection can be described using a BRDF (see \autoref{fig:LIDAR_lte} for an illustration of uniform diffusion around a point, highlighted by the turquoise hemisphere). To accurately simulate the diffusive and reflective properties of surfaces, they are modelled as a collection of small microfacets. The distribution of microfacet orientations is described statistically to account for the roughness, specular and diffusive properties of surfaces.

Perfect diffusion is described by the Lambertian model that relates the intensity of the reflected light to the cosine of its incidence angle, resulting in surfaces (parameterized by brightness $k_R$) that reflect light into all directions equally: 
\begin{align}
    \label{eqn: lambertian}
    f(\mathbf{p}, \omega_o, \omega_i) = \frac{k_R}{\pi}
\end{align}
Besides perfect diffuse reflection, we additionally implement the Oren-Nayar model that represents the microfacet orientation angles by a Gaussian distribution which is parameterized by its standard deviation.
Common LIDAR models assume Lambertian surfaces everywhere. In practice, this assumption is often invalidated -- particularly when mobile robots are employed in indoor areas, such as office spaces or shopping malls, where many highly reflective or transparent surfaces, such as windows, are present.

A bidirectional transmittance distribution function (BTDF) captures the amount of radiance carried by the light transmitted into the reflecting medium (BSDF(2) in \autoref{fig:LIDAR_lte} shows light transmission in glass). Since $F_r(\omega_o)$ is the fraction of light reflected, $1-F_r(\omega_o)$ gives the fraction of light transmitted or refracted. Therefore, by introducing Snell's law of refraction to the integral term in \autoref{eqn: output radiance}, a BTDF $L_0^T$ can be written as
\begin{align}
    \nonumber
    L_0^T(\mathbf{p}, \omega_r) = \frac{\eta_o^2}{\eta_i^2}(1-F_r(\omega_o))L_i(\mathbf{p}, \omega_i),
\end{align}
where $\omega_r$ is the refracted direction of the light ray, $\eta_o$ and $\eta_i$ are the refractive indices of the participating media.

The combination of BTDFs and BRDFs describes the scattering of light upon its encounter with a medium. Any such combination is called a bidirectional scattering distribution function (BSDF). A material is often a combination of multiple BSDF -- for example, transparent plastic, besides diffusing light slightly, has specular reflection and transmission components.


The integral in \autoref{eqn: output radiance} needs to be solved for each light ray which can in most cases only be approximated through sampling. Instead of sampling rays from the light source, we assume the LIDAR is the only source of radiance, and trace rays starting from the sensor's location. Such technique is widely known as Whitted integration~\cite{whitted1980integrator}, a recursive ray-tracing algorithm that generates a tree of light rays (we use a maximum recursion depth of five). At each surface interaction, the raytracing tree is expanded by new rays that are cast into the scene according to the BSDF's defined reflective and refractive properties.

In contrast to the rendering of a typical 2D image, we have to take into account the time it takes for the laser light to be reflected back so that we can simulate the measurement process in a LIDAR. Therefore, for each interaction at point $\mathbf{p}$, we record both the distance travelled so far along the render tree and compute the radiance received from $\mathbf{p}$.
As the light travels a distance $R$, its intensity $I$ attenuates according to the inverse square law $I \propto \frac{1}{R^2}$.


Based on the physics-based rendering toolkit \texttt{pbrt}~\cite{pharr2016pbrt}, we implement ray-tracing with a watertight ray-triangle intersection algorithm and represent all scene geometry as triangle meshes. To efficiently find intersecting meshes, we store the scene geometry in a k-d tree.



\begin{figure}
    \centering
    \includegraphics[width=0.65\columnwidth]{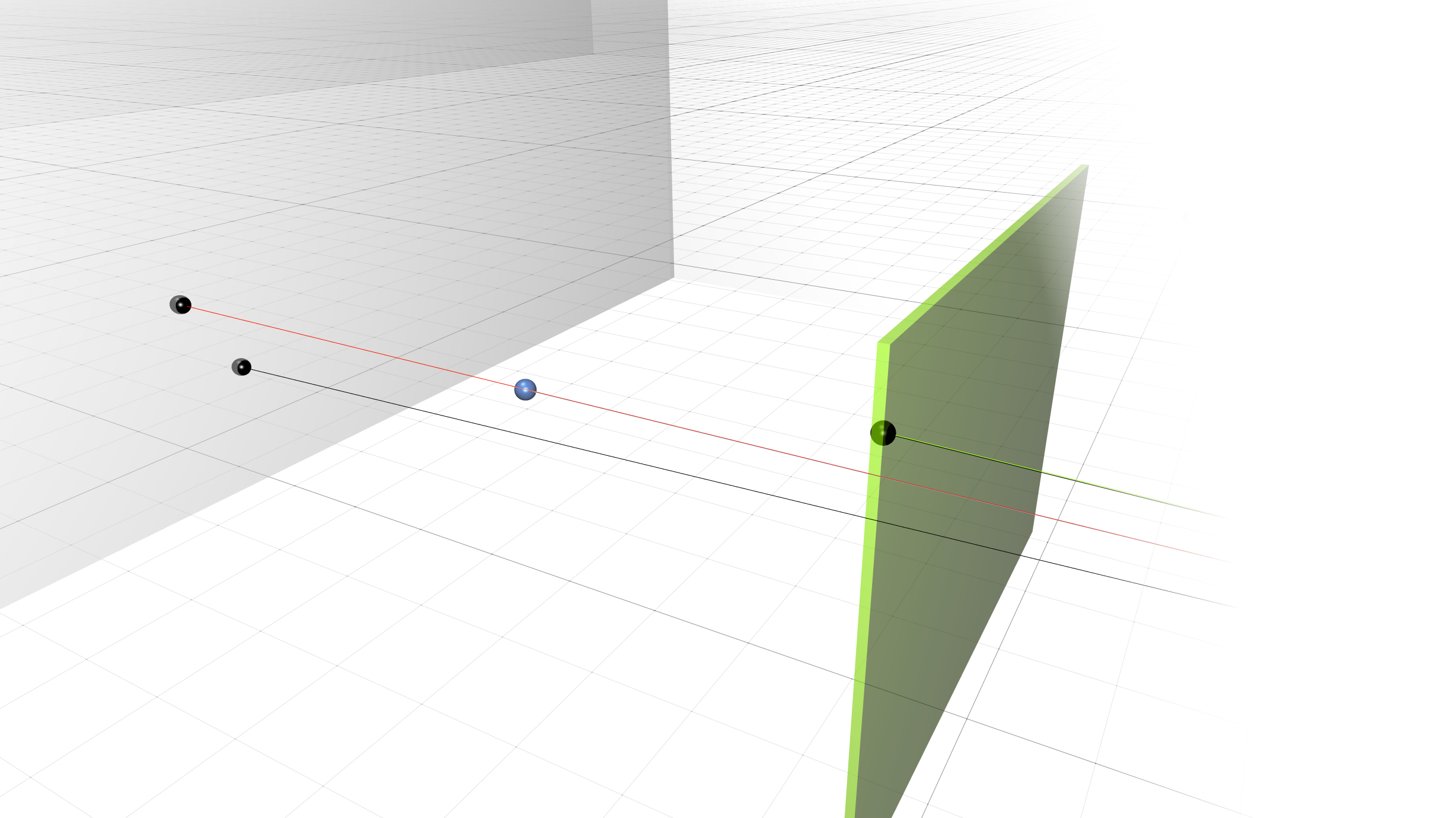}
    \caption{Spurious measurement simulated by sub-sampling the laser beam using three rays (see \autoref{fig:ray_shape} II). Black spheres visualize endpoints of sub-samples, blue point is the final measurement, lying between the green obstacle closer to the LIDAR and the wall further from it.}
    \label{fig:spurious}
\end{figure}

\subsection{Measuring Distance}
\label{sec:distance}

There are two main types of operating principles for LIDARs~\cite{shan2018topographic}: pulsed systems emit a laser pulse and measure the time until the pulse is received to compute the distance. Continuous-wave (CW) LIDARs emit a sinusoidal signal of known wavelength to estimate the distance by measuring the phase difference between the transmitted and received signal (\autoref{fig:LIDAR_lte} right). CW LIDARs typically require less powerful lasers, thus reducing hardware costs.

The LIDAR which we model in this work is the URG-04LX LRF by Hokuyo Automatic Co., Ltd. Being part of the URG series, as described in~\cite{kawata2005urg}, it is a low-cost, planar CW LIDAR with a field-of-view of $240^\circ$, transmitting laser rays into 682 directions (thus achieving an angular resolution of $0.352^\circ$) at a \SI{10}{\Hz} frequency. It modulates the amplitude of the signal to measure distance.

\begin{figure*}[ht!]
    \centering
    \includegraphics[width=.9\textwidth]{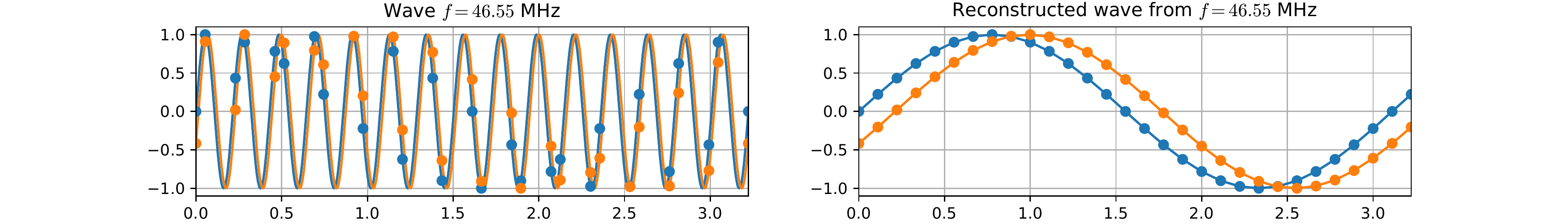}
    \caption{Measurement process in the CW LIDAR URG-04LX \cite{kawata2008urg}. Received (orange) and transmitted (blue) signal at a \SI{46.55}{\MHz} frequency with a phase shift $\Delta\Phi$ resulting from a ranging distance of three meters. \textit{Left:} analogue-to-digital conversion over 15 periods at 30 sampling steps $s_i$ (dots) at times $t_i$ (\autoref{eq:wave_sampling}). \textit{Right:} higher-resolution waveform retained from the coarse samples $s_i$ of the received and transmitted signals.}
    \label{fig:wave_sampling}
\end{figure*}

For a pulse ranging LIDAR, given the two-way travel time $t_R$ of the emitted light, the measured range $\hat{R}$ is computed by
$
\hat{R}=t_R\nicefrac{c}{2},
$
where $c$ is the speed of light. As we are interested in modeling CW LIDARs, the range is computed from the measured phase difference between the transmitted and received signal. $\hat{R}$ is proportional to period time $T=\nicefrac{1}{f}$ and phase difference $\Delta \Phi$:
\begin{align}
\label{eq:distance_phase}
\hat{R} = \frac{c}{2}\frac{T}{2\pi}\Delta\Phi = \frac{1}{4\pi}\frac{c}{f}\Delta\Phi.
\end{align}

In the case of the Hokuyo LRF, the amplitude-modulated laser light is received by an avalanche photodiode~\cite{kawata2005urg}. Throughout the ray-tracing process for a single measurement ray, we compute both the actual range $R^*$ traveled up to the light-surface interaction plus the received radiance $L$ (cf. \autoref{sec:surface}). Each such return produces two new \emph{received} waveforms over 15 periods at $f_1 =\SI{46.55}{\MHz}$ and $f_2=\SI{53.2}{\MHz}$ frequencies with phase shifts $\Delta\Phi^*(f_1)$, $\Delta\Phi^*(f_2) + 90^\circ$,\footnote{Given actual range $R^*$, the actual phase shift for the waveform of frequency $f$ is computed as $\Delta\Phi^*(f) = 4\pi\frac{f}{c}\frac{1}{R^*}$.} and amplitude $\propto L$ (\autoref{fig:LIDAR_lte} right), respectively. For the two frequencies $f_1$ and $f_2$, we sum up the received waveforms until the ray-tracing process is complete.

Next, the measurement process of the Hokuyo LIDAR begins by sampling both waveforms through 30 samples $s_i$ at times $t_i$. The samples are reordered to form a high-resolution waveform (\autoref{fig:wave_sampling} right). Using a simple discrete Fourier series approximation, the phase is computed via
\begin{align}
\label{eq:wave_sampling}
\Phi = \arctan\left({\frac{\sum_i s_i \cos{t_i}}{\sum_i s_i \sin{t_i}}}\right).
\end{align}
Analogously, the phase is computed for the transmitted signal and the phase difference to the received signal is computed by subtracting both phases. From this phase shift $\Delta\Phi$, the measured distance is computed via \autoref{eq:distance_phase}. The two phase shifts from the \SI{46.55}{\MHz} and \SI{53.2}{\MHz} waveforms are compared in order to resolve ambiguity in the phase difference of a single wave~\cite{kawata2005urg}.

\begin{figure}
    \centering
    \includegraphics[width=.85\columnwidth]{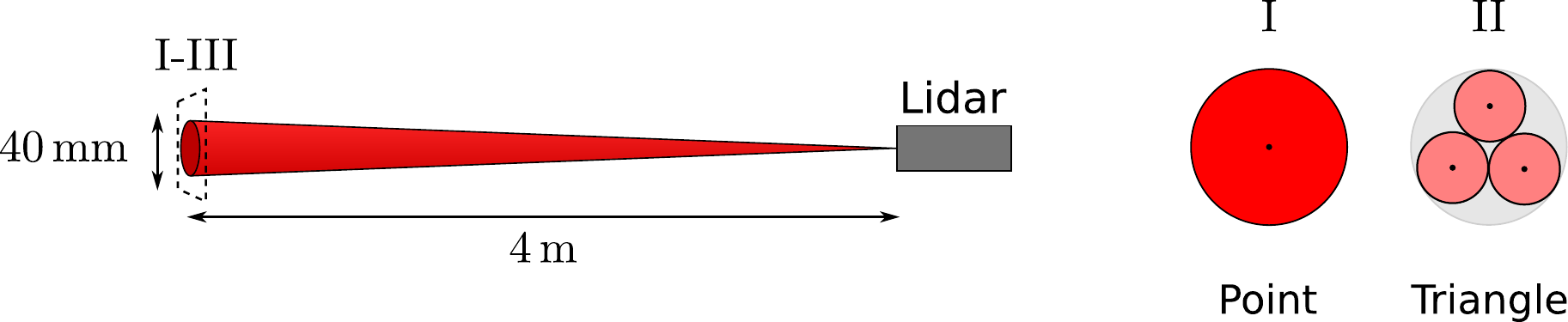}
    \caption{\textit{Left:} The beam divergence of the laser light emitted by the LIDAR amounts to a beam diameter of \SI{4}{\cm} at a distance of \SI{4}{\meter}. \textit{Right:} Cross-section views of approaches to sample the laser beam signals(cone) via rays. Simulating a LIDAR using a single ray (I) does not allow for capturing the beam divergence, while three (II) or more rays offer a more accurate simulation. For three beams, the contribution of each ray is weighted equally so that the total received radiance per beam is the same as for the point under the same conditions.}
    \label{fig:ray_shape}
\end{figure}

LIDARs exhibit spurious measurements (\autoref{fig:spurious}), i.e., when the laser beam partially hits an obstacle closer to the scanner and another obstacle at a greater distance, the measured depth lies somewhere between these two obstacles. This phenomenon follows from the conical shape of the laser beam.
As noted by Rosenberger et al.~\cite{rosenberger2018analysis}, a single infinitesimally thin ray is not sufficient to sample the behavior of beam divergence. Instead, we model the laser beam using three rays which are aligned around the central laser beam ray (\autoref{fig:ray_shape}). At a distance of \SI{4}{\meter}, the beam diameter is specified to be \SI{40}{\mm}.

\section{Experimental Results}
\label{sec:experiments}

\begin{figure*}[ht!]
    \centering
    \includegraphics[height=4cm]{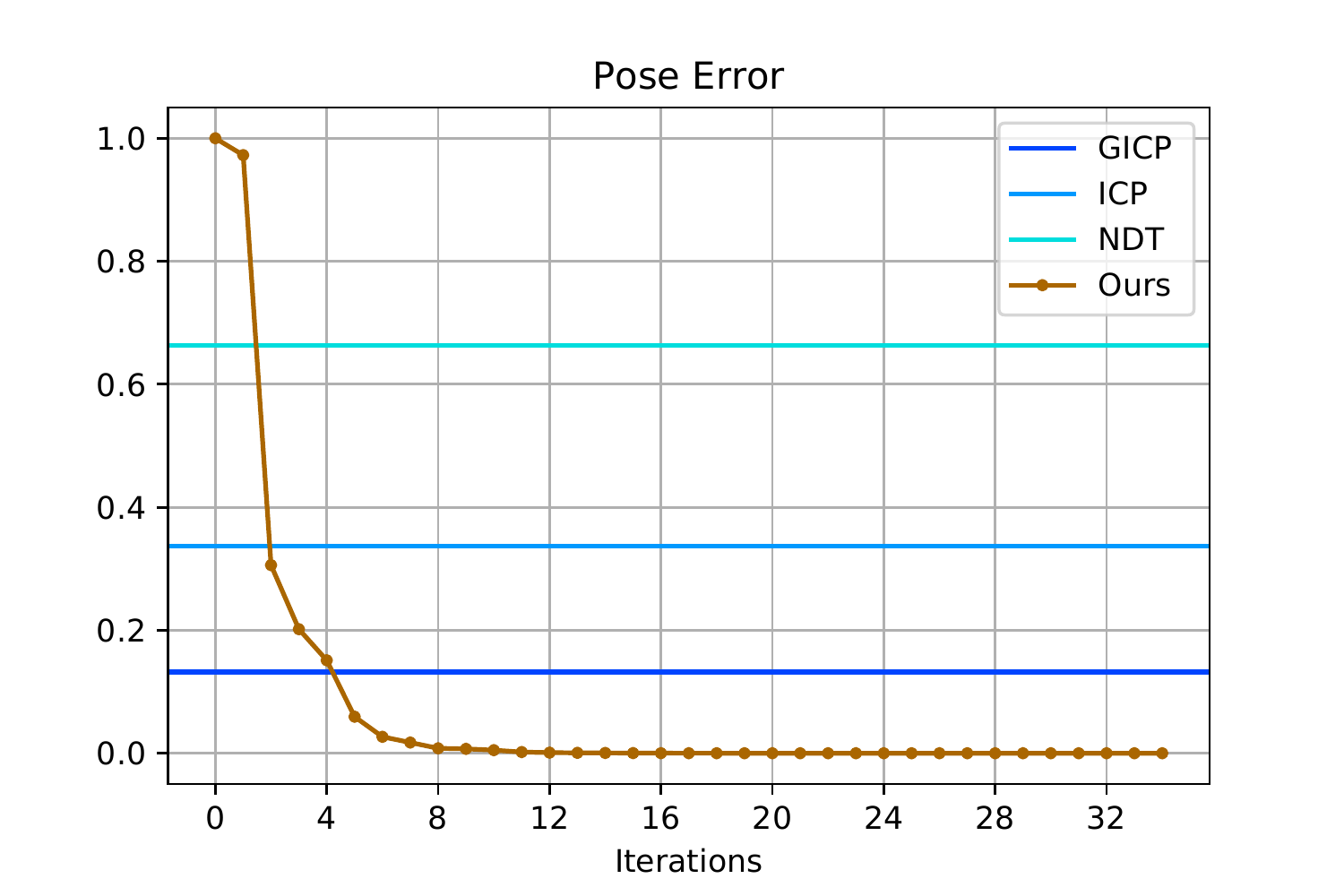}
    \includegraphics[height=4cm,trim=10cm 2cm 10cm 2cm, clip]{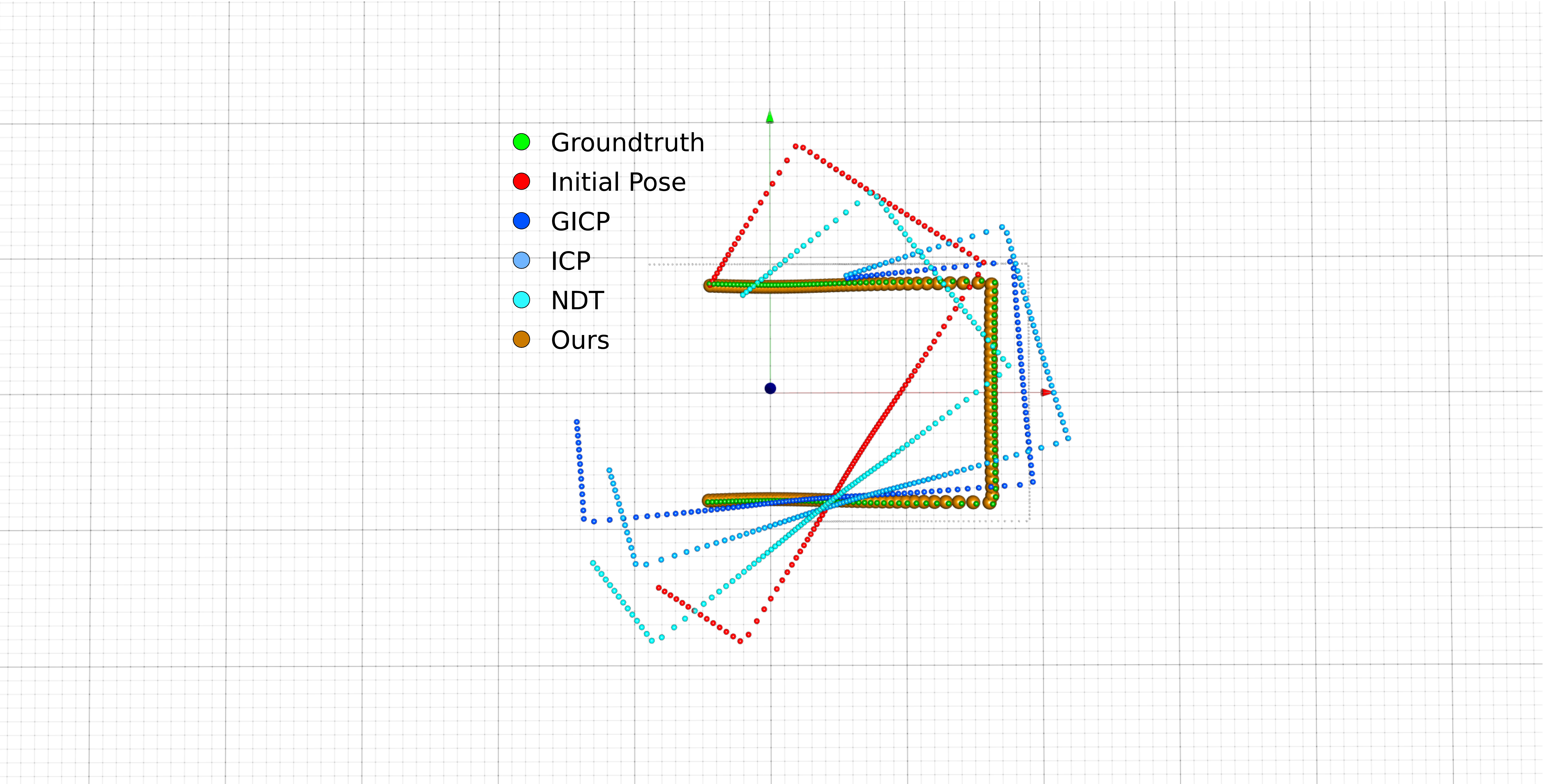}
    \caption{\textit{Left:} Evolution of the pose error, i.e., the SE(2) distance (accounting for yaw angle difference) between the estimated transform and the ground-truth pose of the LIDAR, while localizing the LIDAR in a known environment given its sensor output. The x-axis represents the number of iterations, no iteration-wise information was available for the point cloud registration algorithms GICP, ICP and NDT.
    \textit{Right:} Visualization of the pose estimation results by transforming the point cloud of the given depth measurements by the estimated sensor transform.}
    \label{fig:localization_loss}
\end{figure*}

Besides retaining various effects from the real world, we demonstrate in our experiments the capability of our simulation to serve as a model that can be fitted to actual measurements from a physical LIDAR. To this end, we focus on scenarios where certain inputs to our model need to be estimated -- such as the sensor's pose, material properties of intersecting surfaces, and parameters internal to the sensing process -- so that the resulting simulated measurements closely match the real data.

We investigate several scenarios and replicate them in our simulator to study various applications of the proposed LIDAR model. As shown in \autoref{fig:effects} (3), we created a cuboid environment in which the LIDAR is placed. The environment has the dimensions (width $\times$ depth $\times$ wall height) \SI{185}{\cm} $\times$ \SI{92}{\cm} $\times$ \SI{28}{\cm}. We track the LRF using a Vicon motion capture system and place markers on the objects we want to track to achieve sub-millimeter groundtruth accuracy.

Throughout the experiments, the optimization objective is
\begin{align}
\label{eq:loc_mse}
    \operatorname{minimize}_{\theta} \left(\mathcal{L} = ||f(\theta) - y||^2\right),
\end{align}
where the reality gap between simulated measurements $f(\theta)$ and actual measurements $y$ is to be minimized by adapting the simulation parameters $\theta$. These parameters are defined separately for each experiment.

Our simulator is implemented in C++ and uses the automatic differentiation framework Stan Math~\cite{carpenter2015stan} to algorithmically compute gradients $\frac{\partial\mathcal{L}}{\partial\theta}$. Throughout all experiments, we use an off-the-shelf implementation of the gradient-based optimization algorithm L-BFGS with Wolfe line search from the Ceres library~\cite{ceres-solver} to optimize \autoref{eq:loc_mse} with the calculated gradients throughout all experiments.

\begin{figure*}
    \centering
    \includegraphics[height=4cm, trim=16cm 6cm 18cm 4cm, clip]{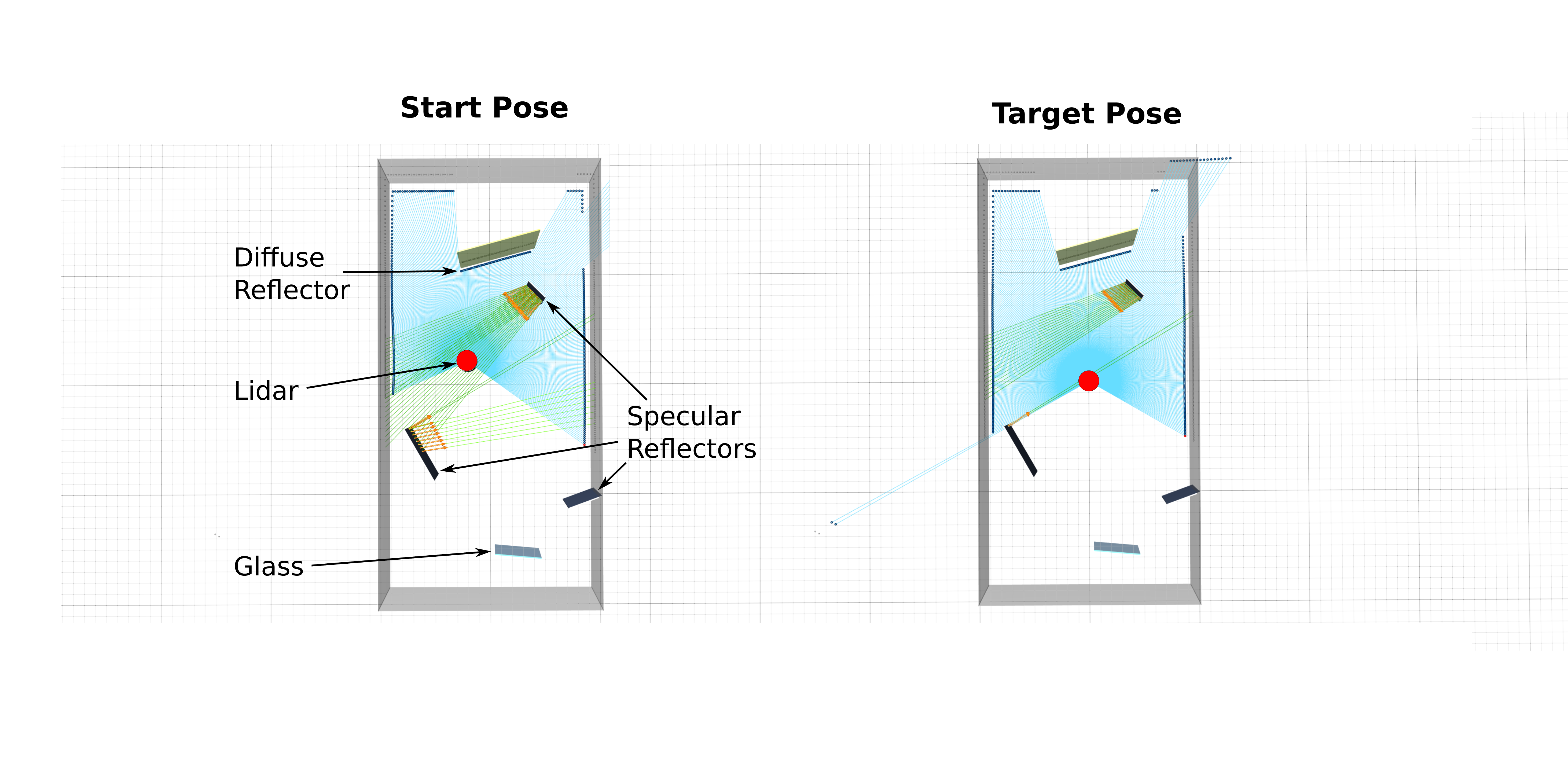}\hspace{1cm}
    \includegraphics[height=4cm]{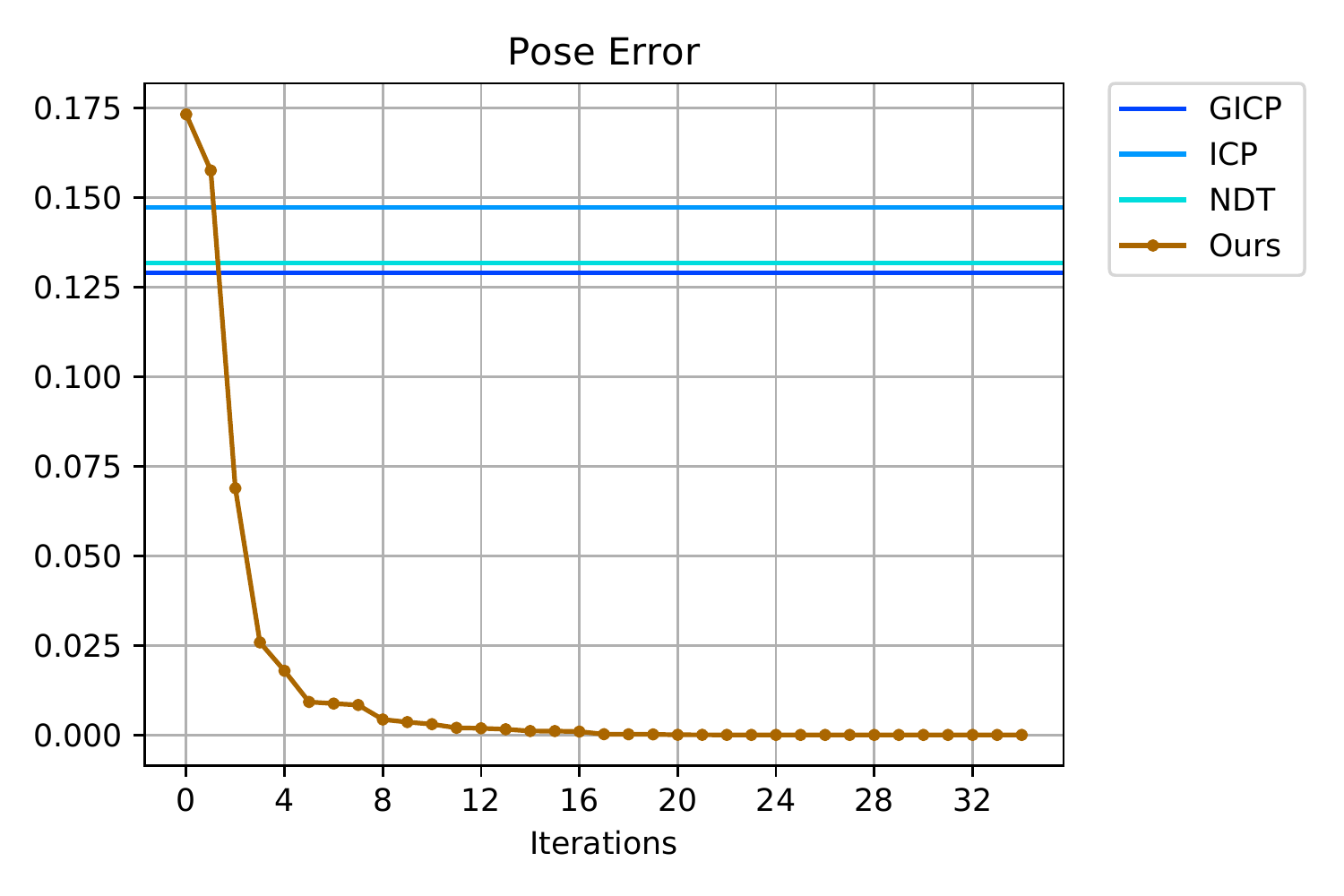}
    \caption{\textit{Left:} Experimental setup of localizing the LIDAR in a known, complex environment consisting of mirrors (specular reflectors), diffuse reflectors (Lambertian) and a glass element. The LIDAR (red) takes measurements (blue dots) at the start pose, and, given the measurements from the target pose, estimate the necessary transform. \textit{Right:} Evolution of the pose error \autoref{eq:loc_mse} (SE(2) distance) between the estimated transform and the groundtruth.}
    \label{fig:complex}
\end{figure*}

\subsection{Localization in a Known Environment}
\label{sec:known_env}

Being placed in a simple cuboid environment where the wall geometries and surface properties are known, the goal in this experiment is to localize the LIDAR. Such a scenario is very common in LIDAR odometry where the current set of range measurements is registered with the point cloud obtained from a previous observation to obtain the rigid transform between the two poses.

Given a set of range measurements, we optimize \autoref{eq:loc_mse} for the sensor's SE(2) pose $\theta$. The initial pose is defined as $(\SI{0}{\meter}, \SI{0}{\meter}, 60^\circ)$, the groundtruth pose is $(\SI{0}{\meter}, \SI{0}{\meter}, 0^\circ)$.
We compare our approach to common point cloud registration algorithms. Given the depth information from the current pose (our initial guess), their task is to find the relative transform to a target point cloud, which stems from the actual LIDAR. Our baselines are Iterative Closest Point (ICP)~\cite{besl1992icp}, Generalized ICP (GICP)~\cite{segal2009gicp}, and Normal Distributions Transform (NDT)~\cite{biber2003ndt}. We use their implementations from the Point Cloud Library~\cite{rusu2011pcl}.

As shown in \autoref{fig:localization_loss} (left), the optimization using our proposed model converges after 34 iterations to the correct pose. Executed on an Intel Core i7-8700K CPU (\SI{3.70}{\giga\Hz}), the computation takes \SI{1.6}{\second}, while the respective point cloud registration algorithms converge after 10-20 iterations (since we do not have access to the iteration-wise performance of these algorithms we show a flat line in the graph). GICP achieves the highest baseline accuracy with a final pose error of 0.13. Although the registration baselines finish the computation by one to two orders of magnitude faster than our approach, we note that our current implementation can be sped up considerably through parallelization since each of the 682 LRF returns can be computed independently.

While classical point cloud registration algorithms are generally able to find good solutions for simple scenarios such as the one shown in \autoref{fig:localization_loss}, where the points follow a simple rectangular geometry, and the shape largely remains constant under the rigid transformation, depth measurements can become more complex when highly reflective surfaces (e.g., mirrors or glass surfaces) are present. Such phenomena become particularly critical in autonomous driving scenarios, where metallic car paint of darker colors absorbs the infrared light from the laser significantly~\cite{seubert2018paint}. Specular reflection and partial transparency in glass windows further distort the obtained depth measurements, affecting many optical sensor pipelines of robots deployed indoors.

To this end, we investigate an entirely simulated scenario where the LIDAR is exposed to multiple specular reflectors (mirrors), a glass surface and a diffuse surface (see \autoref{fig:complex} left). Placed at the initial SE(2) pose $(\SI{0.1}{\meter}, \SI{0.1}{\meter}, 10^\circ)$, the ground truth pose $(\SI{0}{\meter}, \SI{0}{\meter}, 0^\circ)$ needs to be inferred given the sensor measurements. While the initial location is close to the actual pose, the point cloud registration algorithms achieve significantly less improvement over the initial guess compared to our optimization-based approach. As the point clouds differ significantly, a model that relies solely on their shapes is prone to find suboptimal solutions, while the gradient-based optimization applied to \autoref{eq:loc_mse} finds the accurate pose within 34 iterations (see \autoref{fig:complex} right).

\subsection{Tracking a Mirror}
\label{sec:track_mirror}

\begin{figure*}
    \centering
    \includegraphics[height=4cm]{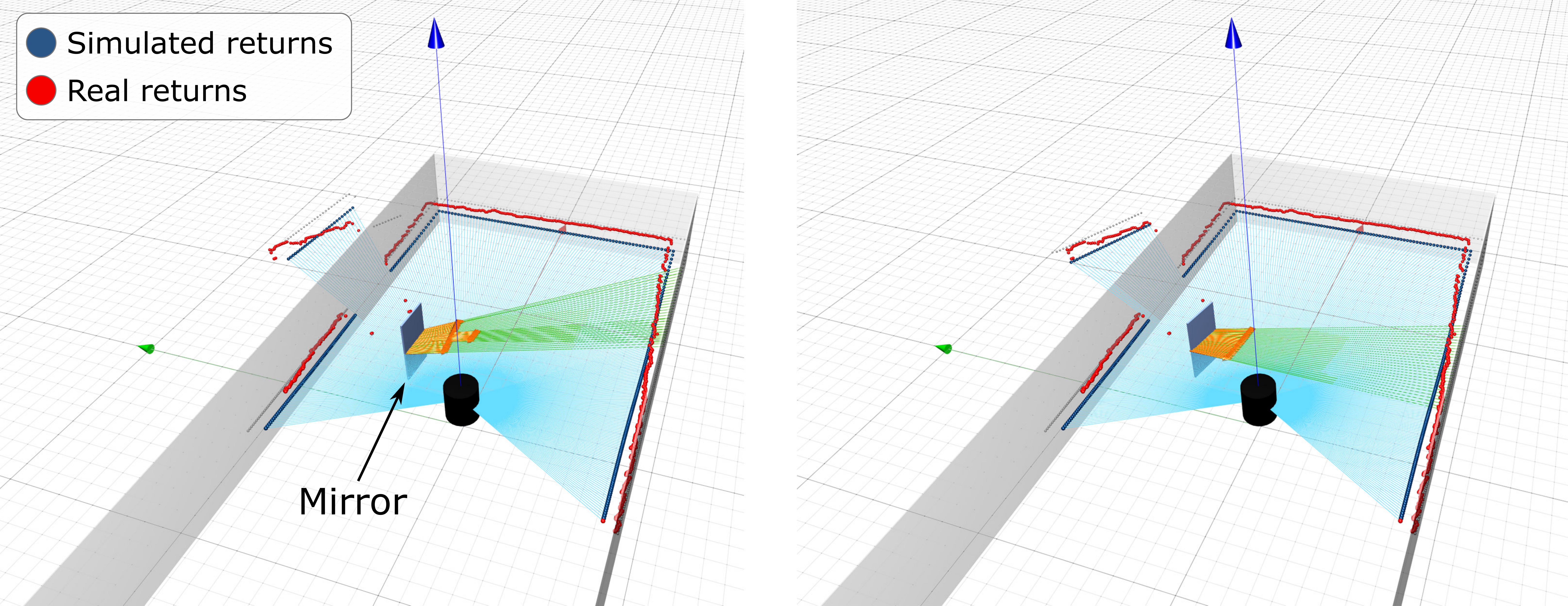}\hspace{.5cm}
    \includegraphics[height=4cm]{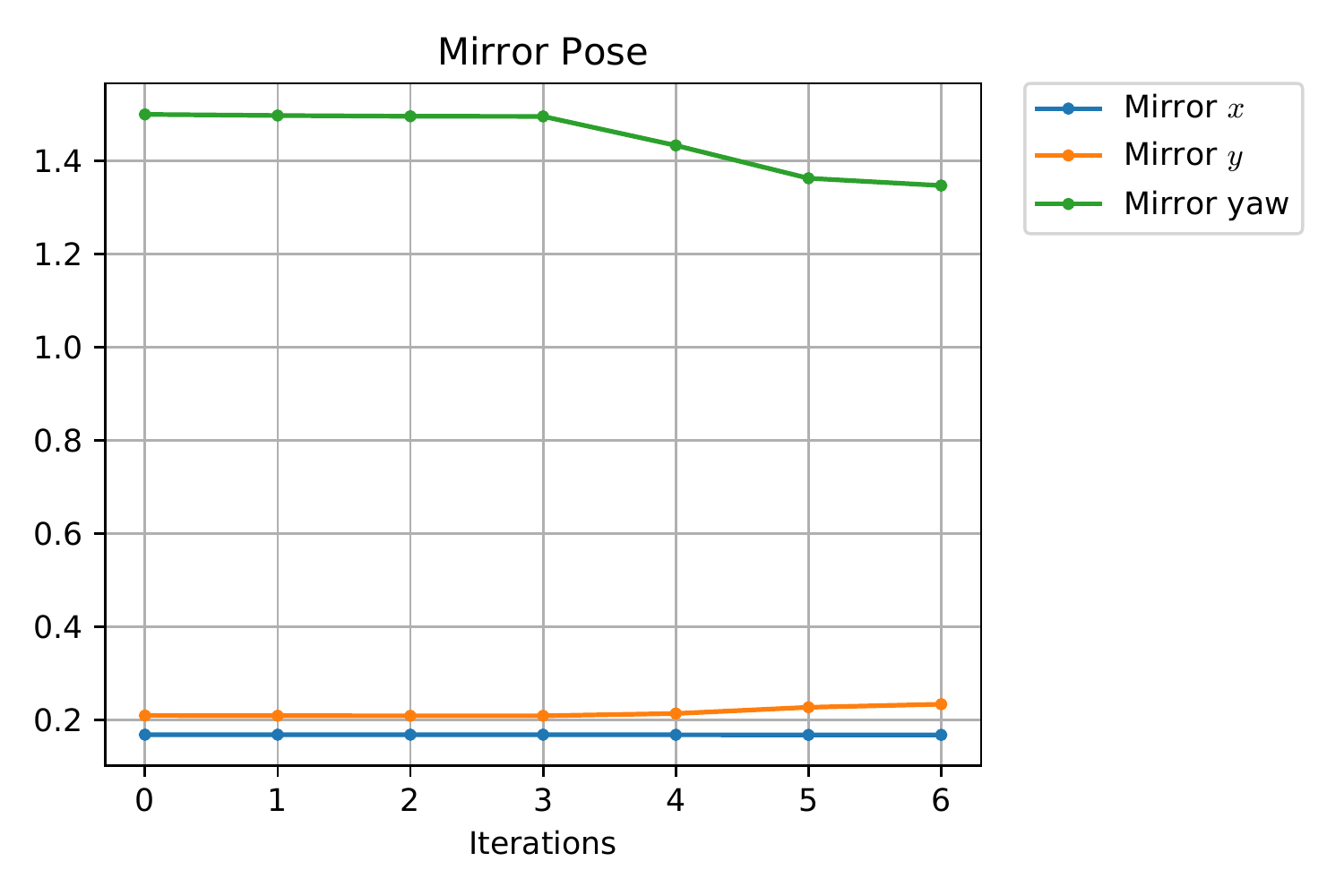}
    \caption{\textit{Left:} Initial setup for the tracking experiment where a mirror needs to be localized in a cuboid environment given real sensor measurements (red dots). Orange arrows with emanating green lines indicate reflection rays causing the measurement rays that hit the mirror to appear further, as the distance to the opposite wall is measured. \textit{Center:} Converged pose estimate of the mirror after 6 iterations. \textit{Right:} Evolution of the mirror's SE(2) pose throughout the optimization.}\vspace{-1em}
    \label{fig:tracking}
\end{figure*}

Besides localizing the LIDAR, other objects in the environment can be tracked if they are modelled within the simulator. As shown in \autoref{fig:tracking}, we place a mirror near the LIDAR and estimate its SE(2) pose using gradient-based optimization. Similar to \autoref{sec:known_env}, our system identification approach applied to our model estimates the correct pose by minimizing the $\ell_2$-norm between the simulated and real depth measurements within six iterations (see \autoref{fig:tracking} right).

Our current approach suffers from the loss of gradient information when the object to be tracked is moved outside the field-of-view of the LIDAR, e.g. when it is behind the walls or in the blind spot of the sensor.
Approaches from the computer graphics community, such as Soft Rasterizer~\cite{liu2019soft}, overcome the issue of missing gradient information by relaxing the intersection checks through a convex combination of blurred edges so that even faces that are hidden contribute to the intersection test.
Future work is directed toward extending our model with such techniques to make tracking and localization more amenable to gradient-based optimization.

\begin{figure*}
    \centering
    \includegraphics[height=4cm, trim=12cm 0 8cm 0, clip]{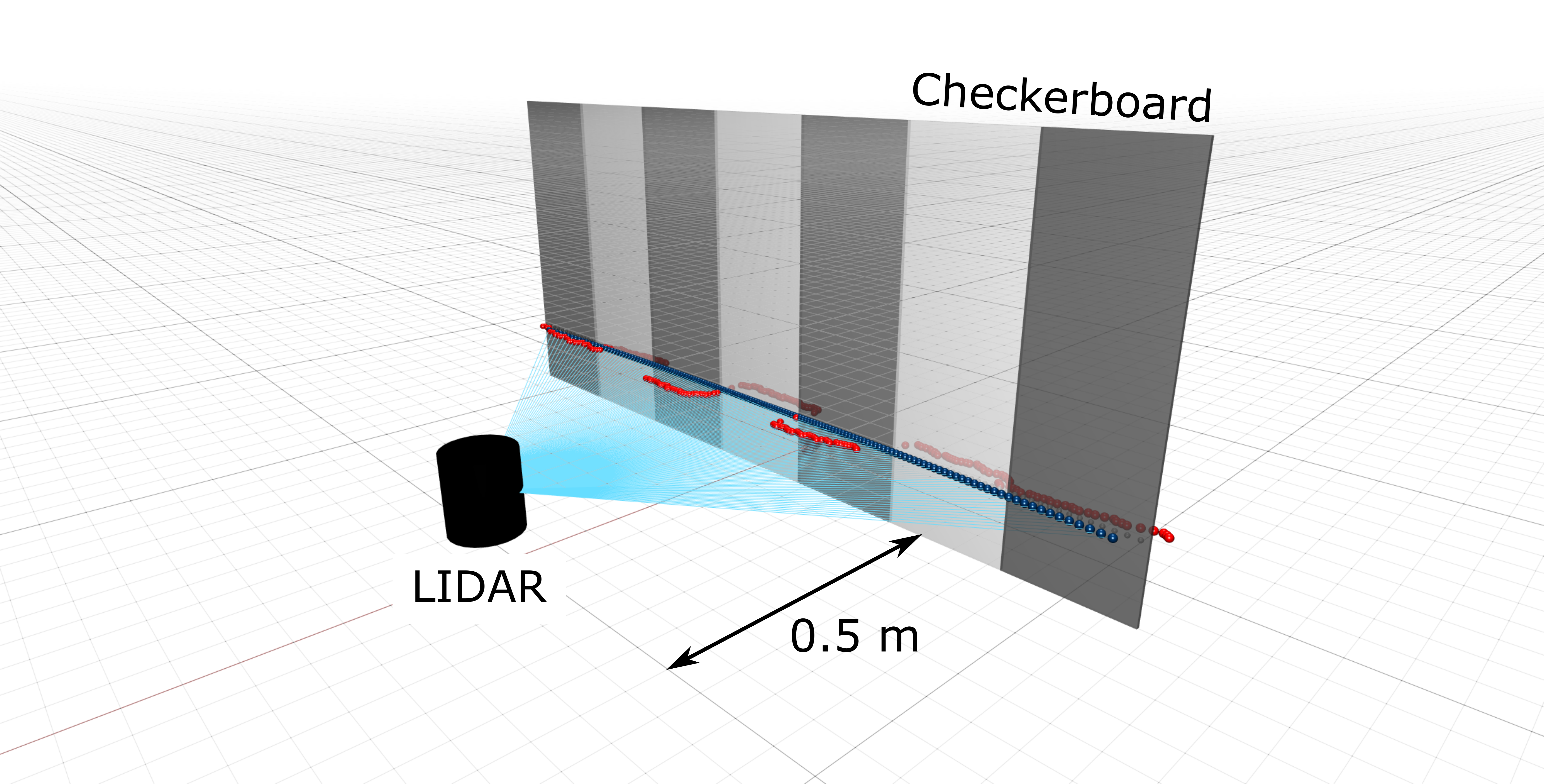}\hspace{.5cm}
    \includegraphics[height=4cm, trim=5cm 0 15cm 0, clip]{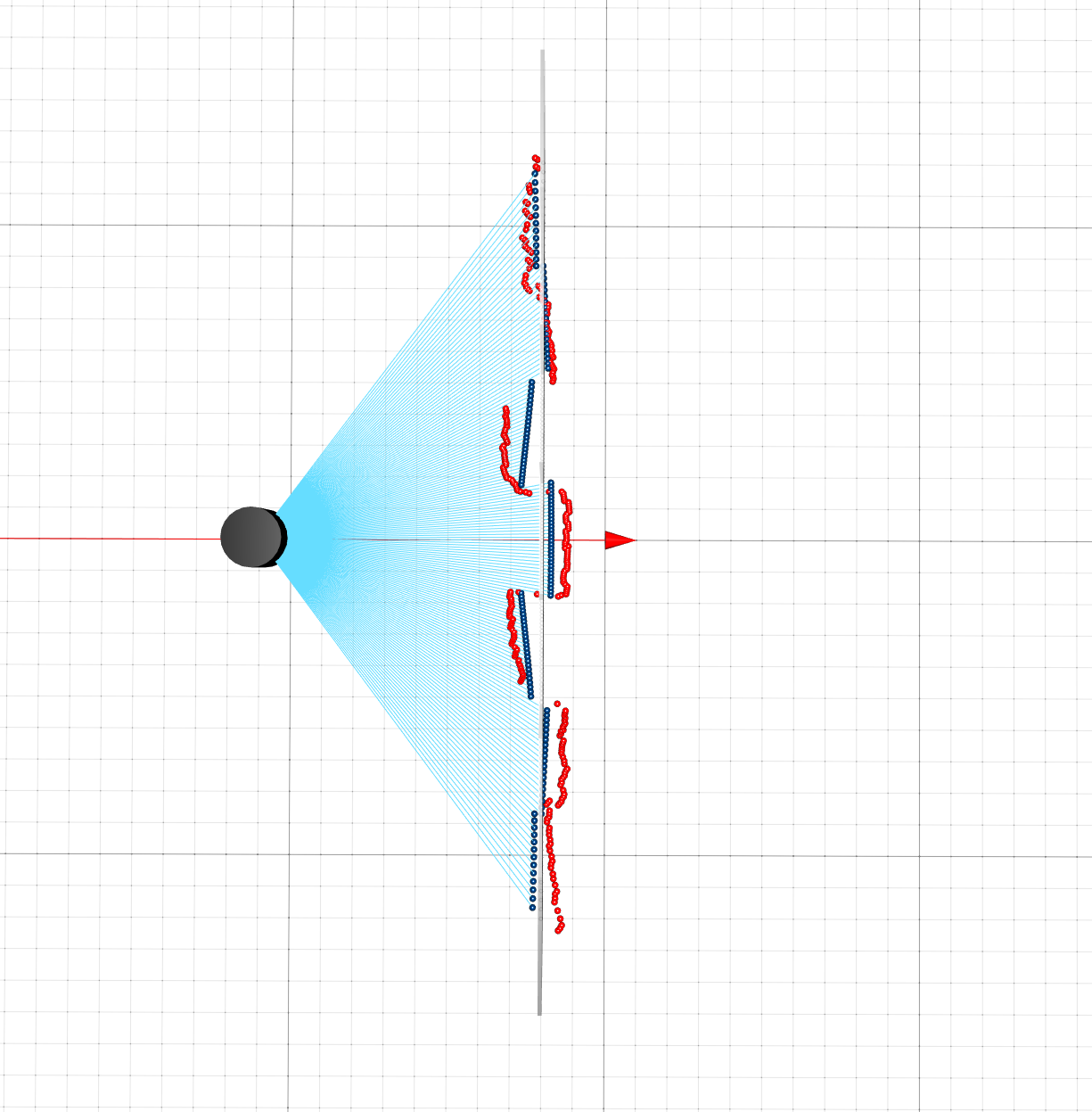}\hspace{.5cm}
    \includegraphics[height=4cm]{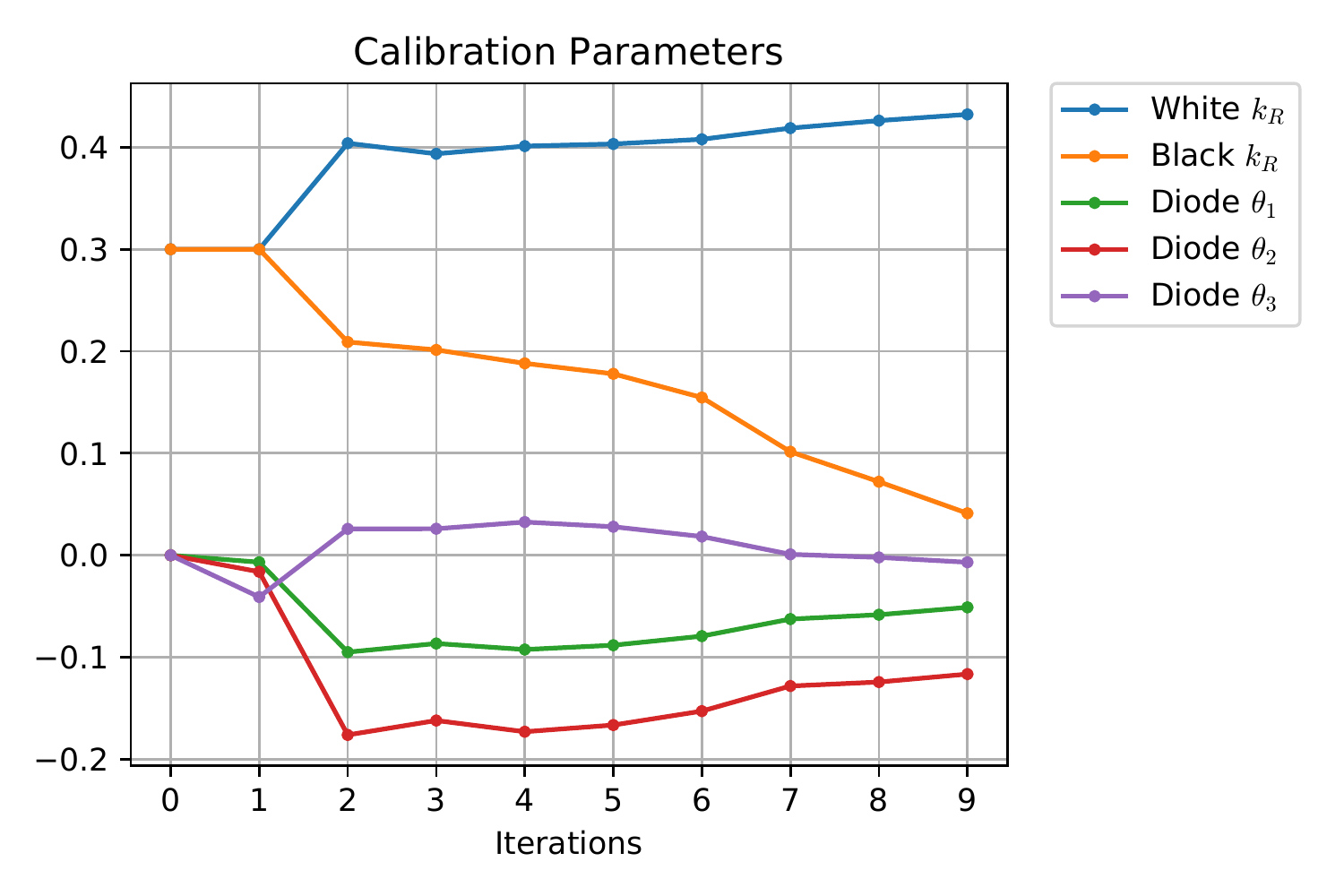}
    \caption{\textit{Left:} Experimental setup of the LIDAR in front of a checkerboard, where measurements are simulated (blue lines visualize rays, blue dots are placed at the simulated depth). Measurements not hitting the checkerboard are not shown. Red dots indicate the measurements from the actual Hokuyo scanner in front of a real checkerboard.
    \textit{Center:} Top-down view of the simulation with converged surface parameters and diode calibration settings. The simulated measurements (blue) appear close to the real measurements (red).
    \textit{Right:} Evolution of the surface properties (Lambertian reflectance coefficients $k_R$ for the white and black sections) and the diode calibration parameters $\theta_1, \theta_2, \theta_3$.}\vspace{-1em}
    \label{fig:calibration}
\end{figure*}

\begin{figure}
    \centering
    \includegraphics[height=3.4cm]{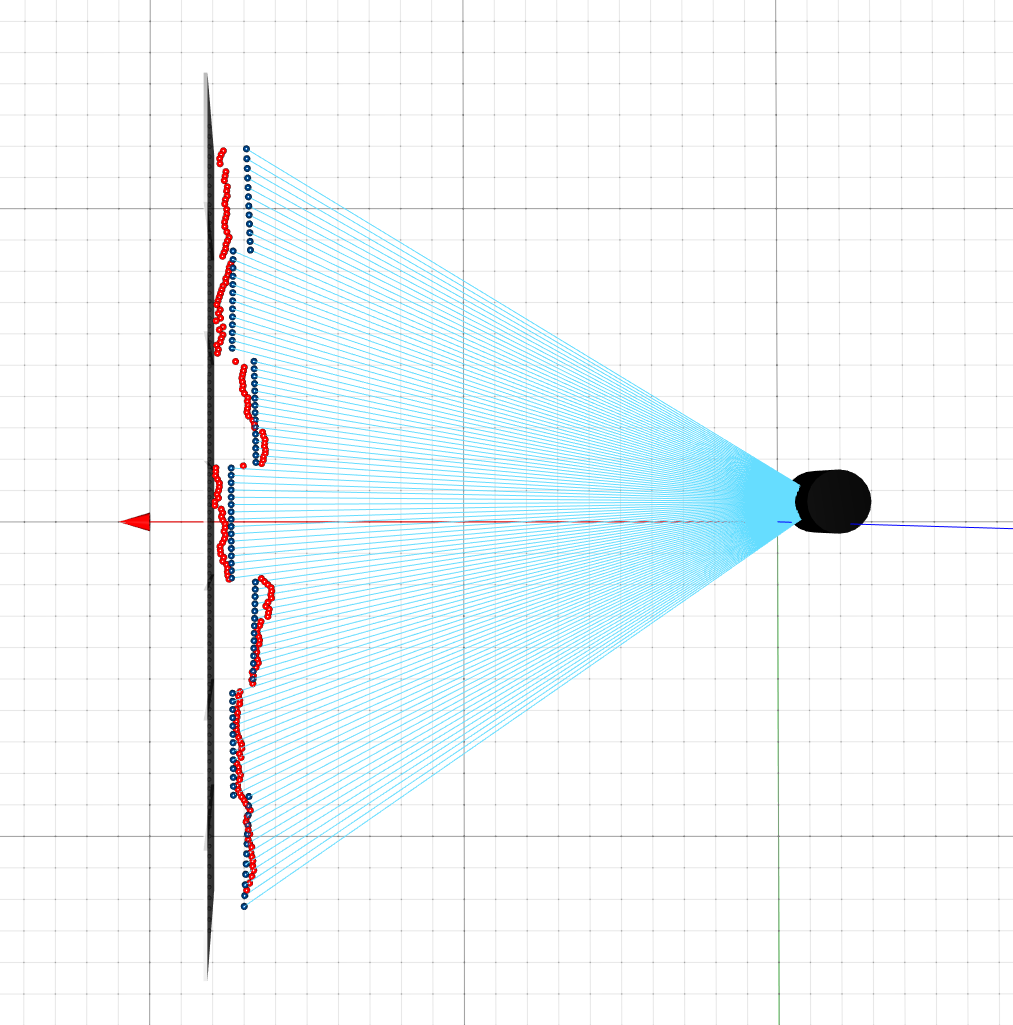}\hspace{1cm}
    \includegraphics[height=3.4cm]{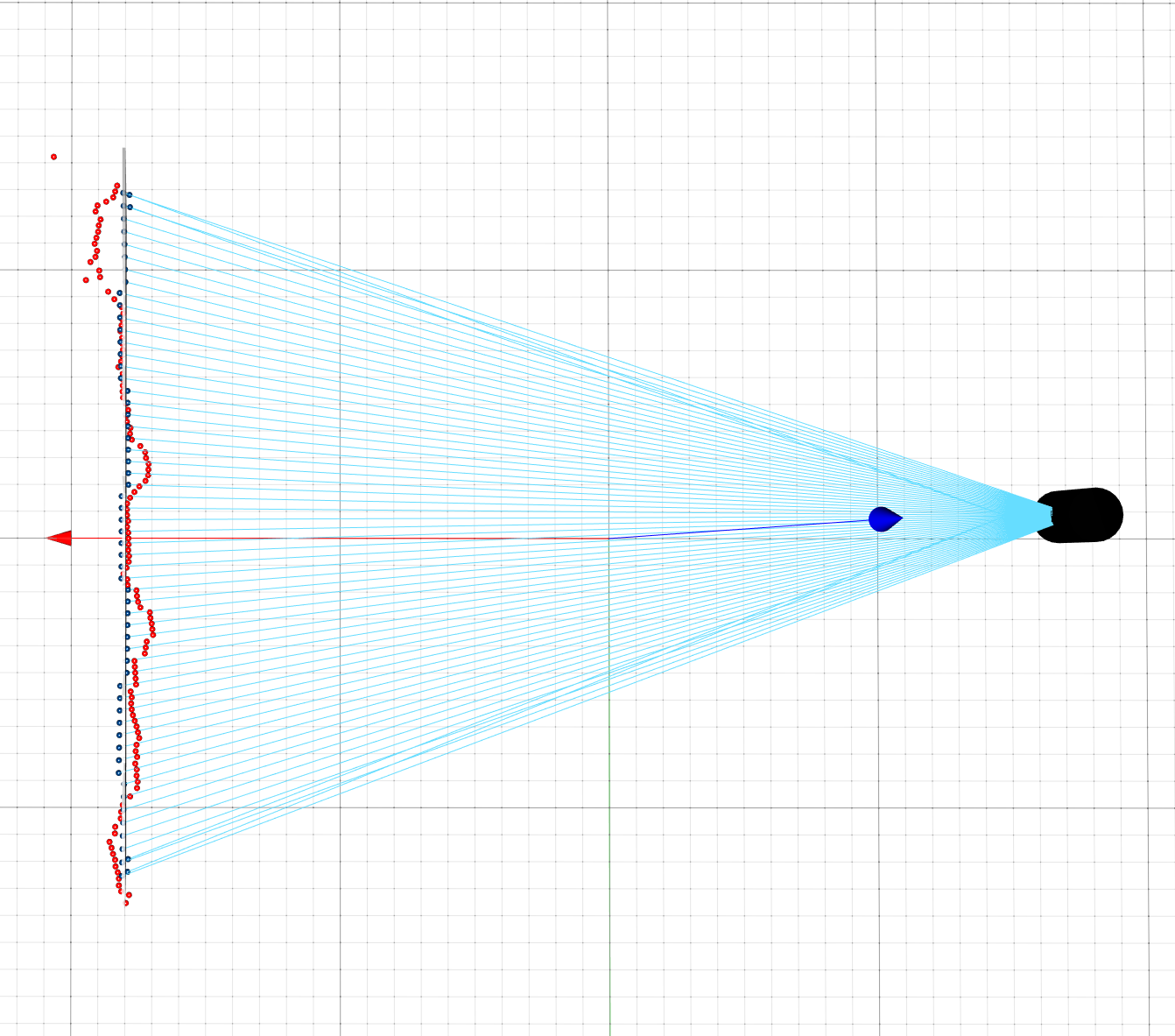}
    \caption{Evaluation of surface properties and diode parameters obtained through the optimization described in \autoref{sec:calibration}. Actual measurements from the real system are shown in red, simulated ranges in blue. \textit{Left:} Checkerboard at a distance of \SI{90}{\cm}. \textit{Right:} Checkerboard at a distance of \SI{1.6}{\meter}.}\vspace{-1em}
    \label{fig:calibration_evaluation}
\end{figure}

\subsection{Diode Calibration and Inferring Surface Properties}
\label{sec:calibration}

As noted in several characterizations of low-energy LIDARs, such as the URG-04LX~\cite{okubo2009urg}, they suffer from a strong dependence of the measured depth on the received radiance. As shown in \autoref{fig:effects} (2), the dark sections of the checkerboard appear closer than the more reflective white sections, resulting in a staircase effect. While the exact cause of this effect is unknown to us, we seek to replicate this behavior through a data-driven approach and leverage our model to infer surface properties given the sensor's range measurements.

We model the relationship between radiance and phase shift as a quadratic polynomial and regress its coefficients through optimizing \autoref{eq:loc_mse}. This can be seen as a form of sum-of-squares optimization \cite{parrilo2012chapter}. The phase shift $\Delta\Phi$ to modulate the two continuous waves in \autoref{eq:distance_phase} is modified as follows:
$
\Delta\Phi^\prime \gets \Delta\Phi - (aL^2 + bL + c),
$
where $L$ is the received radiance and $a,b,c\in\mathbb{R}$ are the polynomial coefficients to be estimated.
We model the checkerboard using dark and bright stripes of the same dimensions as the real checkerboard (\autoref{fig:calibration} left) and assign Lambertian materials of separate reflectivity coefficients $k_R$ (cf. \autoref{eqn: lambertian}) to the black and white sections.

It should be noted that this problem is particularly difficult since we do not have access to the actual intensities of each return but instead only the filtered distance measurements. Nonetheless, we are able to estimate the two different materials of a checkerboard, and converge to polynomial coefficients within 9 iterations (\autoref{fig:calibration} right) that closely match the real measurements (\autoref{fig:calibration} center). When we move the LIDAR from its initial configuration of \SI{0.5}{\meter} distance to other distances in front of the checkerboard, we see that the simulated measurements still retain the staircase pattern from the real world accurately (\autoref{fig:calibration_evaluation}).


\section{Conclusion}
\label{sec:conclusion}

We have presented a physics-based model of a continuous-wave LIDAR that allows for the optimization-based inference of model parameters. It captures the detailed interaction between the laser light and various surfaces, accounting for geometrical and material properties. Through experiments with real-world measurements from the Hokuyo URG-04LX LRF, we have demonstrated various applications of our model in localization, calibration and tracking.


Future research is directed towards extending our proposed simulation to model more powerful pulsed LIDAR sensors and investigate its application in real-world domains on a larger scale.






\bibliographystyle{IEEEtran}
\bibliography{literature}  

\end{document}